\documentclass[10pt,twocolumn,letterpaper]{article}

\usepackage{wacv}
\usepackage{times}
\usepackage{epsfig}
\usepackage{amssymb}
\usepackage{latexsym}
\usepackage{graphicx}
\usepackage{amsmath}
\usepackage{bbm}
\usepackage{color}
\usepackage{adjustbox}
\usepackage{multirow}

\wacvfinalcopy % *** Uncomment this line for the final submission

\ifwacvfinal
\def\assignedStartPage{1} % *** Enter the assigned starting page number (instead of 9876)
\fi

\ifwacvfinal
\usepackage[breaklinks=true,bookmarks=false]{hyperref}
\else
\usepackage[pagebackref=true,breaklinks=true,colorlinks,bookmarks=false]{hyperref}
\fi

\ifwacvfinal
\else
\fi

\begin{document}

%%%%%%%%% TITLE
\title{Inferring Point Clouds from Single Monocular Images by Depth Intermediation}

\author{Wei Zeng\\
University of Amsterdam\\
{\tt\small w.zeng@uva.nl}
\and
Sezer Karaoglu\\
3DUniversum\\
{\tt\small s.karaoglu@3duniversum.com}
\and
Theo Gevers\\
University of Amsterdam\\
{\tt\small th.gevers@uva.nl}
}

\maketitle
%\thispagestyle{empty}

%%%%%%%%% ABSTRACT
\begin{abstract}
   In this paper, we propose a pipeline to generate 3D point cloud of an object from a single-view $RGB$ image. Most previous work predict the 3D point coordinates from single $RGB$ images directly. We decompose this problem into depth estimation from single images and point cloud completion from partial point clouds. 
   
   Our method sequentially predicts the depth maps from images and then infers the complete 3D object point clouds based on the predicted partial point clouds. We explicitly impose the camera model geometrical constraint in our pipeline and enforce the alignment of the generated point clouds and estimated depth maps. 
   
   Experimental results for the single image 3D object reconstruction task show that the proposed method outperforms existing state-of-the-art methods. Both the qualitative and quantitative results demonstrate the generality and suitability of our method.
\end{abstract}

%%%%%%%%% BODY TEXT
\section{Introduction}
\label{Introduction}

Inferring 3D shapes from 2D images is an important computer vision task which has many applications such as robot-environment interaction, 3D-based classification and recognition, virtual and augmented reality. Recently, due to the development of deep learning techniques and the creation of large-scale datasets \cite{chang2015shapenet}, increasing attention has been focused on deep 3D shape generation from single $RGB$ images \cite{choy20163d, hane2017hierarchical, tatarchenko2017octree, jiang2018gal, wang2018pixel2mesh, groueix2018atlasnet, mescheder2019occupancy, nguyen2019graphx}.

A number of previous methods represent the estimated 3D shape as a voxelized 3D occupancy grid \cite{choy20163d, hane2017hierarchical, gwak2017weakly, yang20173d, tatarchenko2017octree, richter2018matryoshka, mescheder2019occupancy}. While it may seem straightforward to extend 2D CNNs to process 3D data by utilizing 3D convolutional kernels, data sparsity and computational complexity are the restrictive factors of this type of approaches. The source of data sparsity is that most of the information, which is needed to compute the 3D structure, is provided by the surface voxels. In fact, the part which the shape representation lies on the surface of the 3D object, makes up only a small fraction of all voxels in the occupancy grid. This makes 3D CNNs computational expensive yielding considerable amount of overhead during training and inference. To overcome these issues, recent methods focus on designing neural network architectures and loss functions to process and predict 3D point clouds. These point clouds consist of points which are uniformly sampled over the object surfaces. For example, Fan \cite{fan2017point} introduces a framework and loss functions designed to generate unordered point clouds directly from 2D images. Jiang \cite{jiang2018gal} extends this pipeline by adding geometrically driven loss functions for training. Groueix \cite{groueix2018atlasnet} represents a 3D shape as a collection of parametric surface elements to infer the surface representation of the shape. However, the inference procedure does not explicitly impose any geometrical constraint. Therefore, these models purely rely on the quality of training data and the effectiveness of learning to generalize.

%%%%%%%

\begin{figure}[t]
\includegraphics[width=0.45\textwidth]{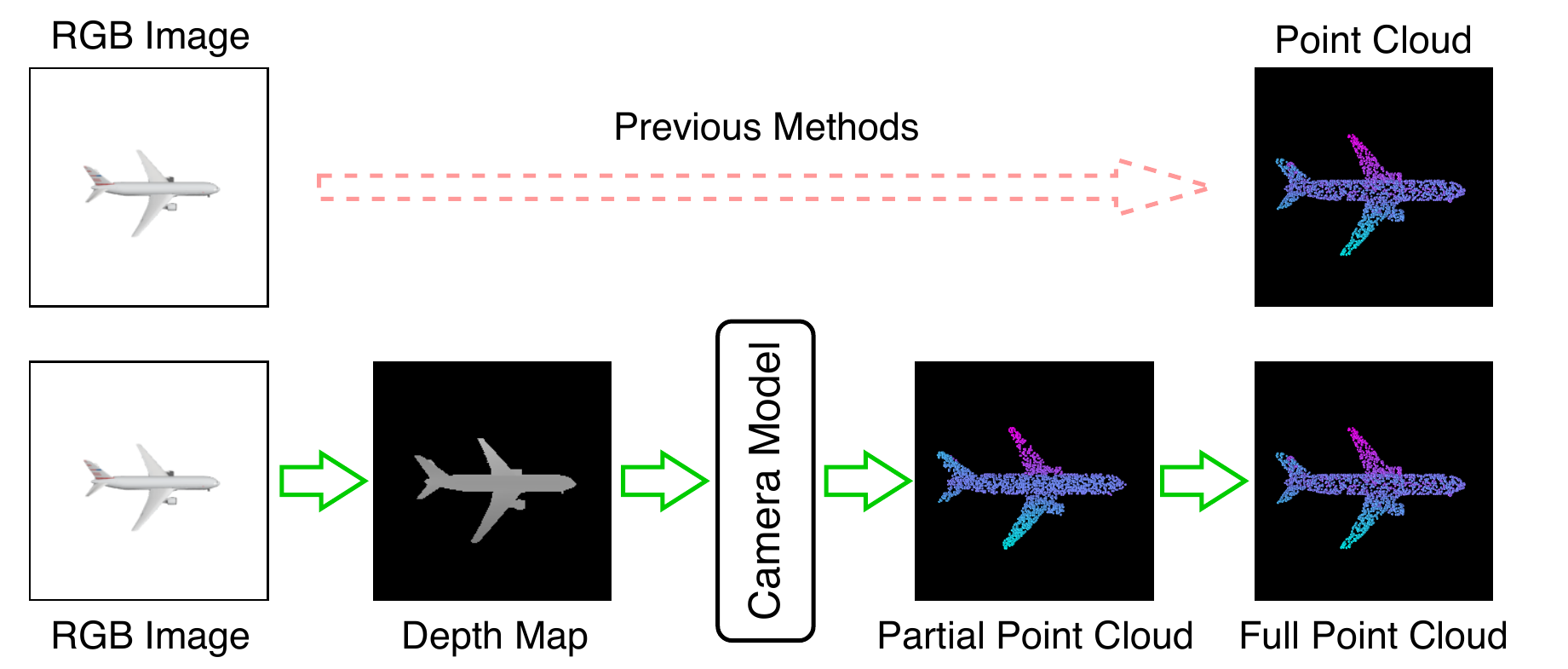}
\centering
\caption{Most of the existing methods generate point clouds directly from $RGB$ input images. In contrast, our method first predicts the depth map of the $RGB$ input image and infers the partial (view-specific) point cloud. The transformation of the partial point cloud is based on the camera model. In this way, the camera model is explicitly used as a geometrical constraint to steer the 2D-3D domain transfer. Then, a full 3D point cloud is generated. A 3D-2D refinement process is used to enforce the alignment between the generated full 3D point cloud and the depth map prediction.}
\label{fig:idea}
\end{figure}

%%%%%%%

%%%%%%%

\begin{figure*}[t]
\includegraphics[width=\textwidth]{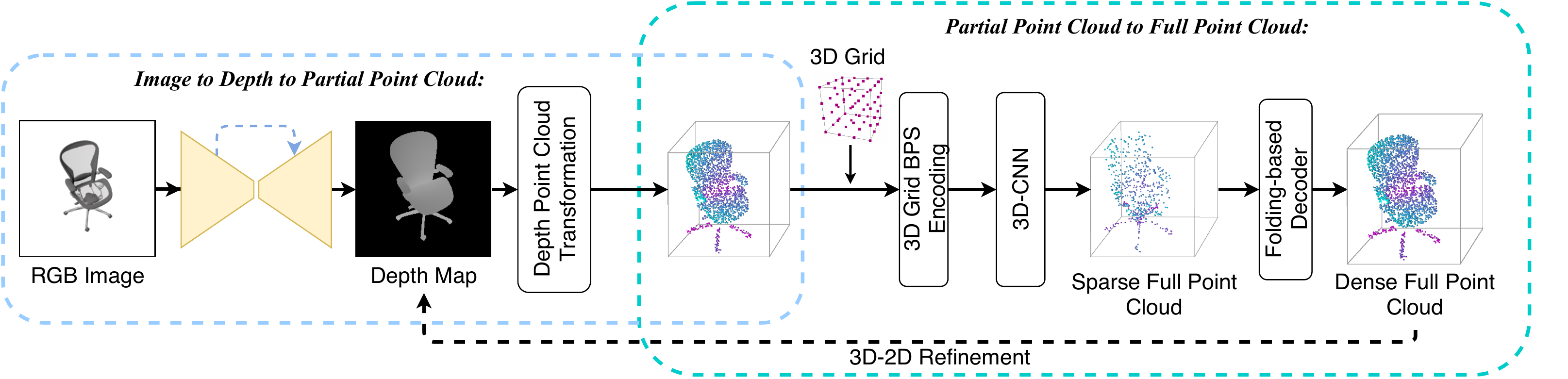}
\centering
\caption{Overview of our framework. Our proposed network receives a $RGB$ image as input. It predicts the depth map of the input image, and calculates the partial point cloud based on the camera geometry. Then, the predicted partial point cloud is encoded by the unit 3D grid basis point set and taken by a 3D convolutional neural network to produce the sparse full point cloud. The final full point cloud is generated in a sparse-to-dense fashion via a folding-based decoder. Finally, the 3D-2D refinement module enforces the alignment between the generated full 3D point cloud and the estimated depth map.}
\label{fig:architecture}
\end{figure*}

%%%%%%%

In this paper, we propose a pipeline to sequentially predict the depth map to infer the full 3D object shape, see Fig.~\ref{fig:idea}. The transformation of the depth map into the partial point cloud is driven by the camera model. In this way, the camera model is explicitly used as a geometrical constraint to steer the 2D-3D domain transfer. Our method is composed of three components, namely, depth intermediation, point cloud completion and 3D-2D refinement, see Fig.~\ref{fig:architecture} for a detailed overview of our pipeline. 

First, given a single $RGB$ image of an object, the depth intermediation module predicts the depth map, and then computes the point cloud of the visible part of the object in image space. We refer to this (single-view) point cloud as the partial point cloud. The computation of the partial point cloud is based on the camera model geometry. In this way, we explicitly impose the camera model as a geometrical constraint in our transformation to regulate the 2D-3D domain transfer. 

Then, the point cloud completion module infers the full point cloud using the partial point cloud as input. To preserve the context of point clouds and utilize neighboring relationships between points, partial point clouds are first encoded by unit 3D grid basis point sets. Then, a 3D convolutional neural network is used to compute context-aware features. The output is further processed by a folding-based decoder to generate a full point cloud.

Finally, the 3D-2D refinement process enforces the alignment between the generated full point cloud and the depth map prediction. The refinement module imposes a 2D projection criterion on the generated point cloud together with the 3D supervision on the depth estimation. This self-supervised mechanism enables our network to jointly optimize both the depth intermediation and the point cloud completion modules. 

In summary, our contributions are as follows:

\begin{itemize}
    \item A novel neural network pipeline to generate 3D shapes from single monocular $RGB$ images by depth intermediation. 
    \item Incorporating the camera model as a geometrical constraint to regulate the 2D-3D domain transfer.
    \item A 3D-grid based point cloud completion module to generate fine-grained full point clouds.
    \item A 3D-2D refinement module to jointly optimize both depth estimation and point cloud generation.
    \item Superior performances on the task of 3D single-view reconstruction on both synthetic dataset (ShapeNet) and real dataset (Pix3D) to demonstrate the generality and suitability of the proposed method.
\end{itemize}
%-------------------------------------------------------------------------

%%%%%%%%%%%%%%%%%%%%%%%%%%%%%%%%%%%%%%%%%%%%%%%%%%%%%%%%%%%%%%%%%%%%%%%

\section{Related Work}

\textbf{Depth Estimation} Single-view, or monocular, depth estimation refers to the problem where only a single image is available at test time. Eigen \cite{eigen0} shows that it is possible to produce pixel-wise depth estimation using a two scale deep network which is trained on images with their corresponding depth values. Several methods extend this approach by introducing new components such as CRFs to increase the accuracy \cite{depth_crf}, changing the loss from regression to classification \cite{depth_classification}, using other more robust loss functions \cite{fcrn}, and by incorporating scene priors \cite{depth_surface}. Recently, there are a number of methods to estimate the depth in an unsupervised way. Godard \cite{depth_unsupervised_1} proposes an unsupervised deep learning framework by introducing loss functions which impose consistency between predicted depth maps which are obtained from different camera viewpoints. Kuznietsov \cite{depth_unsupervised_2} adopts a semi-supervised deep learning method to predict depth maps from single images. As opposed to existing methods, in our work, we use supervised depth estimation to produce depth maps to enable the inference of 3D shapes. Moreover, our 3D-2D refinement module uses the generated full point cloud as a 3D supervision algorithm to steer the depth estimation.

\textbf{Feature Learning on Point Clouds} Because of the irregular nature of point clouds, they cannot be processed in a straightforward manner by standard grid-based CNNs. Only recently, a number of methods are proposed that apply deep learning directly on (raw) 3D point clouds. PointNet \cite{pointnet} is the pioneering work that directly processes 3D point sets in a deep learning setting. The modified version of PointNet, PointNet++ \cite{pointnet++}, abstracts local patterns by sampling representative points and recursively applying PointNet as a learning component to obtain the final representation.  Zeng \cite{wei_3dcontextnet} introduces 3DContextNet that exploits both local and global contextual cues imposed by the k-d tree to learn point cloud features hierarchically. Yang \cite{yang2018foldingnet} proposes a folding-based decoder that deforms a canonical 2D grid onto the underlying 3D object surface of a point cloud. Prokudin \cite{prokudin2019efficient} introduces basis points sets to obtain a compact fixed-length representation of point clouds. In this paper, we leverage regular 3D grids as basis point sets to regularize unordered partial point clouds. In this way, the network is able to learn spatial-context aware features to complete the missing parts of the partial point clouds.

\textbf{3D Shape Completion} Shape completion is an essential task in geometry and shape processing. The aim of conventional methods is to complete shapes using local surface primitives, or to formulate it as an optimization problem \cite{nealen2006laplacian, sorkine2004least}. With the advances of large-scale shape repositories like ShapeNet \cite{chang2015shapenet}, researchers start to develop fully data-driven methods. For example, 3D ShapeNets \cite{wu20153d} use a deep belief network to obtain a generative model for a given shape database. Nguyen \cite{thanh2016field} extends this method for mesh repairing. Most of the existing learning-based methods represent shapes by voxels. In contrast, our method uses point clouds. Point clouds preserve the full geometric information about the shapes while being memory efficient. Related to our work is PCN \cite{yuan2018pcn}, which uses an encoder-decoder network to generate full point clouds in a coarse-to-fine fashion. However, the proposed method is not limited to the shape completion task. Our aim is to generate the full point cloud of an object from a single $RGB$ image.

\textbf{Single-image 3D Reconstruction} Traditional 3D reconstruction methods are, in general, based on multi-view geometry. The major research directions include structure from motion (SfM) \cite{sfm} and simultaneous localization and mapping (SLAM) \cite{slam}. Recently, increasing attention has focused on data-driven 3D voxel reconstruction from single images \cite{choy20163d, fan2017point, yang20173d}. Choy \cite{choy20163d} proposes 3D-R2N2. The method takes as input one or more images of an object taken from different viewpoints. The output is the reconstruction of the object in the form of a 3D occupancy grid by means of recurrent neural networks. As a follow-up work, Gwak \cite{gwak2017weakly} makes use of foreground masks for 3D reconstruction by constraining the reconstruction to be in the space of unlabeled real 3D shapes. Wu \cite{wu2017marrnet} also attempts to reconstruct the 3D shapes from 2.5D sketches. They first compute the 2.5D sketches of objects and then treat the predicted 2.5D sketches as intermediate images to regress the 3D shapes. Tulsiani \cite{tulsiani2018multi} presents a framework that allows to learn a single view prediction of a 3D structure without direct supervision of shape or pose. Richter \cite{richter2018matryoshka} poses 3D shape reconstruction as a 2D prediction problem to leverage well-proven architectures for 2D pixel-prediction. Mescheder \cite{mescheder2019occupancy} implicitly represents the 3D surface as the continuous decision boundary of a deep neural network classifier. Different from the above methods, our proposed approach explicitly imposes the camera model in the 2D-3D transformation and infers the partial point clouds from predicted depth maps purely based on 3D geometry. 

Voxel-based methods are computationally expensive and are only suitable for coarse 3D voxel resolutions. To overcome this issue, Fan \cite{fan2017point} introduces a framework to regress unordered point clouds directly from 2D images. Jiang \cite{jiang2018gal} extends this pipeline by adding geometrically driven loss functions for training. Groueix \cite{groueix2018atlasnet} introduces an approach to generate parametric surface elements for 3D shapes. The learnable parametrizations transform a set of 2D squares to the surface, covering it in a way similar to an atlas. Mandikal \cite{mandikal20183d} proposes a latent-embedding matching method to learn the prior over 3D point clouds. It first trains a 3D point cloud auto-encoder and then learns a mapping from the 2D image to the corresponding learned embedding. Wang \cite{wang2018pixel2mesh} represents 3D meshes in a graph-based convolutional neural network and produce correct geometry by progressively deforming an ellipsoid, leveraging perceptual features extracted from the input image. Nguyen \cite{nguyen2019graphx} proposes to blend the image features with a random point cloud and deform it to the final representative point set of the object. Different from these above methods, our approach sequentially predicts the depth map, infers the partial point cloud based on the camera model, and generates the full point cloud of the 3D shape. In addition, the proposed method explicitly enforce the alignment between the generated point cloud and the estimated depth map by jointly optimizing both of the components. 

%The relevant work to our method is GenRe proposed by Zhang et al \cite{genRe}.  Both methods factorize $f_{2D \to 3D}$ into geometry projections and learnable reconstruction modules, the differences are as follows: \emph{(1)} Shape completion space. Our method performs shape completion in a \emph{3D point-cloud space}, while~\cite{genRe} performs spherical map inpainting in a \emph{2D image space}. \emph{(2)} End-to-end training. Our method is \emph{fully differentiable} and can be trained \emph{end-to-end}, while~\cite{genRe} is not. To project depth to a spherical map,~\cite{genRe} casts rays from each UV coordinate on the unit sphere to the center of the sphere to generate the spherical representation. This process part is not differentiable. In contrast, our method converts depth maps to point clouds using camera parameters. Our process is fully differentiable. So our pipeline can be trained end-to-end and jointly optimize both the depth intermediation and the point cloud completion modules. \emph{(3) } Efficiency of the model. Our model has one projection from depth maps to point clouds and performs 2D convolutions on point coordinates. In contrast,~\cite{genRe} has three geometry projections and perform 3D convolutions on voxels. Compared with~\cite{genRe}, our model is faster in inference time (\textbf{51ms} vs. 542ms) with a smaller model size (\textbf{180MB} vs. 452MB).

%-------------------------------------------------------------------------

%%%%%
\begin{figure}[t]
\includegraphics[width=0.45\textwidth]{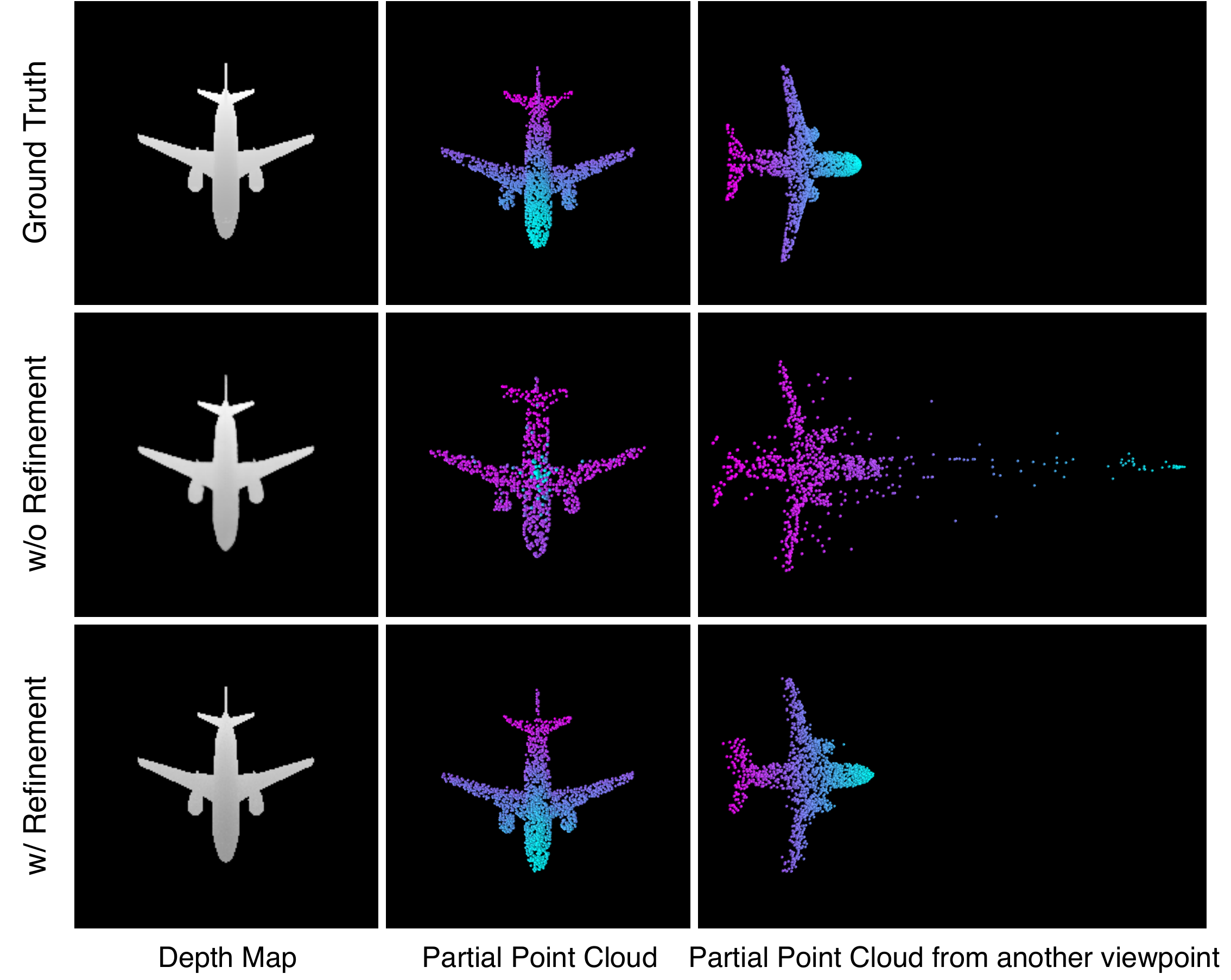}
\centering
\caption{Depth maps and their corresponding partial point clouds. From top to bottom: (1) ground truth, (2) depth estimation without and (3) depth estimation with 3D-2D refinement. It can be (visually) derived that when depth estimation is transformed into a partial point cloud (based on the camera model), the predicted partial point cloud without refinement may suffer from errors (i.e. "flying" points). This is clearly visible in the second row. This type of estimation errors are largely reduced by our 3D-2D refinement process (third row). Best viewed in color.}
\label{fig:partial point cloud}
\end{figure}

%%%%%

\section{Method}

We propose a pipeline that generates point clouds from $RGB$ images by depth intermediation. To compute a 3D point cloud from a single-view $RGB$ image, our network uses three modules: (1) a depth intermediation module is proposed to predict depth maps and calculate the partial point clouds based on the camera model geometry; (2) a point cloud completion module is proposed to infer full 3D point clouds from predicted partial point clouds; (3) and a 3D-2D refinement mechanism is proposed to enforce the alignment between the generated point clouds and the estimated depth maps. Our full pipeline can be trained in an end-to-end fashion and enables to jointly optimize both depth estimation and point cloud generation.

\subsection{Depth Intermediation}

The first component of our network takes a 2D $RGB$ image of an object as input. It predicts the depth map of the object and calculates the (visible) point cloud based on the camera model. The aim of the depth intermediation module is to regulate the 2D-3D domain transfer and to constrain the structure of the learned manifold. Most of the previous methods directly generate the 3D shape from a single 2D image. Although they use geometry-driven loss functions during training, the inference procedure does not explicitly impose any geometrical constraint. In contrast, our method uses the predicted depth map to compute the partial point cloud. In this way, during inference, geometrical constraints are still explicitly incorporated by means of depth estimation and the camera model. 

An encoder-decoder network architecture is used for our depth estimation. Note that any deeper depth estimation networks can be easily plugged in our proposed pipeline, due to the simplicity of the object-level depth estimation, we stay with the simple configuration of the architecture in this work. The encoder is a VGG-16 \cite{vgg} architecture up to layer conv5\underline{ }\underline{ }3 encoding a $224\times224$ $RGB$ image into 512 feature maps of size $7\times7$. The decoder contains five $3\times3$ deconvolutional layers with layer sizes (256, 128, 64, 64, 64). Then, four $1\times1$ convolutional layers with layer sizes (64, 64, 64, 1) are applied to encourage individuality to the generated pixels. Skip connections link the related layers between the encoder and decoder. The output is the corresponding depth map with the same resolution as the 2D $RGB$ input image. 

Then, the partial point cloud is computed using the camera model. For a perspective camera model, the correspondence between a 3D point $(X,Y,Z)$ and its projected pixel location $(u,v)$ on an image plane is given by:
\begin{equation} \label{eq:1}
    Z[u,v,1]^T=\textbf{K}(\textbf{R}[X,Y,Z]^T +\textbf{t})
\end{equation}
where $\textbf{K}$ is the camera intrinsic matrix. $\textbf{R}$ and $\textbf{t}$ denote the rotation matrix and the translation vector, which are already included because the partial point cloud is view-specific. So in this work, it simplifies to $Z[u,v,1]^T=\textbf{K}[X,Y,Z]^T$.
We assume that the principal points coincide with the image center, and that the focal lengths are known. Note that when the exact focal length is not available, an estimation (approximation) may still suffice. %When the object is reasonably distant from the camera, larger focal lengths will choose between perspective and weak-perspective models.

In general, object-level depth estimation is coarse. Hence, the corresponding partial point cloud may suffer from noise (e.g. flying points) at the boundaries along the frustum. The aim of our 3D-2D refinement is to enforce the partial point cloud to be consistent with the full point cloud. The goal is to reduce the estimation errors at the boundaries. For example, consider Fig.~\ref{fig:partial point cloud}, where depth maps and their corresponding partial point clouds are shown. The predicted partial point cloud without refinement (second row) suffers from errors (i.e. flying points). This type of estimation errors are largely reduced by our 3D-2D refinement process (third row).

\subsection{Point Cloud Completion}

The full point cloud is inferred by learning a mapping from the space of partial observations to the space of complete shapes. Most previous methods (e.g. PCN \cite{yuan2018pcn} and FoldingNet \cite{yang2018foldingnet}) use Multi-layer Perceptrons (MLPs) to directly process point clouds, which may cause the loss of details because the structure and context of point clouds are not fully considered. Inspired by basis point sets (BPS) \cite{prokudin2019efficient}, in this paper we encode the partial point clouds as minimum distances to a fixed set of 3D grid points. Having the partial point cloud $X=\{x_1,...x_n, x_i  \in R^3\}$ and the unit 3D grid basis point set $B=\{b_1,...b_k, b_j \in R^3\}$ (in this work we use $k=32^3$), we compute the directional delta vector from each basis point to the nearest point in the partial point cloud:
\begin{equation} \label{eq:bps}
    X^B = \{(\mathop{argmin}\limits_{x_i \in X}d(b_j, x_i)- b_j)\} \in R^{k \times 3}
\end{equation}
In this way, the structure and context of point clouds are explicitly preserved by the 3D grid representation. Furthermore, encoding by the 3D grid basis point set regularizes the unordered partial point cloud which allows the network to fully utilize the neighborhood relationship to learn context-aware features.

\begin{figure}[t]
\includegraphics[width=0.45\textwidth]{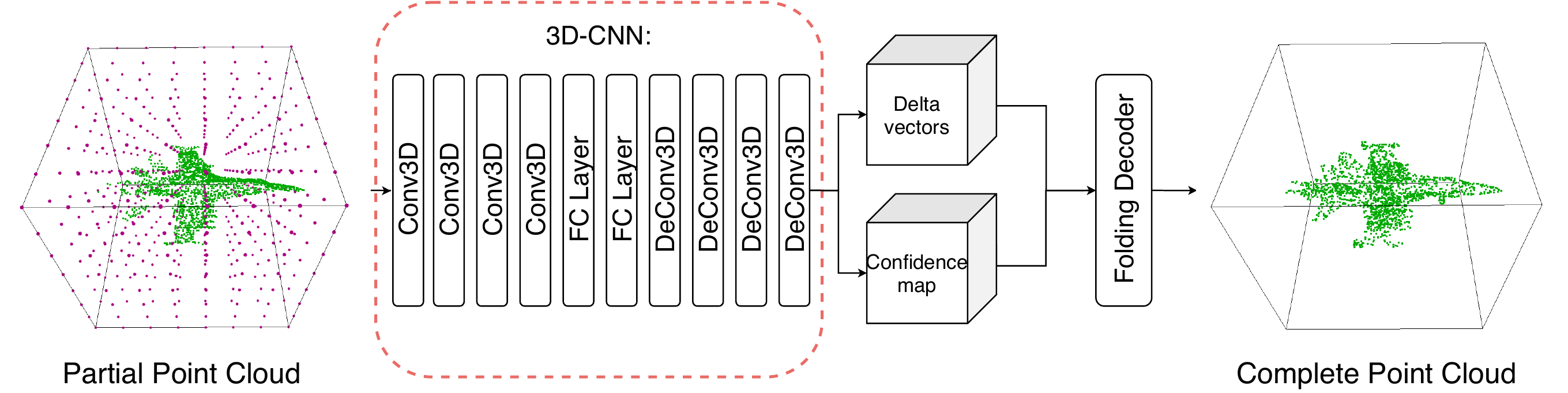}
\centering
\caption{Detail architecture for the encoder part of the point cloud auto-encoder. The encoder is composed of PointNet layers and graph-based max-pooling layers. }
\label{fig:encoder_part}
\end{figure}

As shown in Fig.~\ref{fig:encoder_part}, after encoded by the 3D grid basis point set, a 3D convolutional neural network (3D-CNN) is applied to complete the missing parts of the partial point cloud. The encoder of the 3D-CNN has four 3D convolutional layers with layer size (32, 64, 128, 256), each of the layers is followed by a max-pooling layer with a kernel size of $2^3$. The encoder is then followed by two fully connected layers with sizes of 1024 and 2048. The decoder consists of four deconvolutional layers to transpose back to the original size of the 3D grid basis point set. The output of the 3D-CNN has two branches: one is the set of predicted delta vectors of the complete point cloud with respect to the 3D grid. The other is the confidence map for each point of the 3D grid. The confidence map represents the distances from the basis points to the nearest points in the complete point cloud, with higher confidence indicating the closer distance between points. Then, the point coordinates are recovered by adding the output delta vectors to the 3D grid basis point set. We sub-sample $m=256$ sparse point clouds according to the confidence map as key point sets to abstract the entire 3D shapes and input them to the following folding-based decoder to generate the final full point cloud. For each key point $\hat{x}_i$, a patch of $t=u^2$ points ($u=2$ in our experiments) is generated in local coordinates centered at $\hat{x}_i$ via the folding-based decoder. Eventually, a $N=1024$ complete point cloud is generated as output of the network. Note that here we set $N=1024$ in order to have a fair comparison with existing methods as it is the common choice.

%%%%%

\subsection{3D-2D Refinement}

In this section, the aim is (1) to align the predicted point cloud and the corresponding estimated depth map and (2) to jointly optimize both the depth intermediation and the point cloud completion module. 

For the depth intermediation network, flying points may occur in the inferred partial point cloud near the object boundaries along the frustum, as shown in Fig.~\ref{fig:partial point cloud}. The cause of this is the lack of contextual information for object-level depth estimation. Therefore, the aim of the 3D-2D refinement is to reduce these estimation errors (i.e. depth noise reduction). 

To reduce the depth estimation errors, the generated point cloud is used as a 3D self-supervision component. A point-wise 3D Euclidean distance is used between the partial point cloud and the full point cloud, which is defined by: 
\begin{equation} \label{eq:2}
    L_{d}(P_p,P_f)=
    \sum_{p_i\in P_p}\underset{p_j\in P_f}{min}\|p_i-p_j\|_2^2 
\end{equation}
where $P_p$ and $P_f$ are the predicted partial point cloud and the predicted full point cloud, respectively. This regularizes the partial point cloud to be consistent with the full point cloud with the aim to reduce the noise.

To constrain the generated point cloud using the 2D projection supervision, we penalize points in the (full point cloud) projected image $I_p$ which are outside the silhouette $I_s$: 
\begin{equation} \label{eq:3}
    L_p = \sum_{q_i\in Q_p} 
    \mathbbm{1}((I_{p}(q_i)-I_{s}(q_i))>0)
    \underset{q_j\in Q_s}{min}\|q_i-q_j\|_2^2
\end{equation}
where $Q_p$ and $Q_s$ represent the pixel coordinates of the projected image and the silhouette, respectively. $\mathbbm{1}(.)$ is an indicator function set to 1 when a projected point is outside the silhouette. The goal of this constraint is to recover the details of the 3D shape.

\subsection{Discussion}

The relevant work to our method is GenRe proposed by Zhang et al \cite{genRe}.  Both methods factorize $f_{2D \to 3D}$ into geometry projections and learnable reconstruction modules, the differences are as follows: \emph{(1)} Shape completion space. Our method performs shape completion in a \emph{3D point-cloud space}, while GenRe performs spherical map inpainting in a \emph{2D image space}. \emph{(2)} End-to-end training. Our method is \emph{fully differentiable} and can be trained \emph{end-to-end}, while GenRe is not. To project depth to a spherical map, GenRe casts rays from each UV coordinate on the unit sphere to the center of the sphere to generate the spherical representation. This process part is not differentiable. In contrast, our method converts depth maps to point clouds using camera parameters. Our process is fully differentiable. So our pipeline can be trained end-to-end and jointly optimize both the depth intermediation and the point cloud completion modules. \emph{(3) } Efficiency of the model. Our model has one projection from depth maps to point clouds and performs 2D convolutions on point coordinates. In contrast, GenRe has three geometry projections and perform 3D convolutions on voxels. Compared with GenRe, our model is faster in inference time (\textbf{51ms} vs. 542ms) with a smaller model size (\textbf{180MB} vs. 452MB). GenRe generalizes well to diverse novel objects from categories not seen during training, but for the single image 3D object reconstruction task, experimental results in the next section show that the proposed method outperforms GenRe for all of the 13 categories in ShapeNet dataset. 
%Furthermore, experimental results in the next section show that the proposed method outperforms GenRe on the single image 3D object reconstruction task. 

%-------------------------------------------------------------------------

%%%%
\begin{table*}[t]
\caption{Quantitative comparison of Chamfer Distance and Earth Mover's Distance metric on ShapeNet. Our proposed method outperforms the state-of-the-art for most of the categories and achieves a lower overall mean error in both CD and EMD metrics}
\label{table:overall}
\begin{center}
\begin{adjustbox}{width=\textwidth}
\begin{tabular}{c | c | c  c  c  c  c  c  c  c  c  c  c  c  c | c }
\hline
   & & airplane & bench & cabinet & car & chair & monitor & lamp & speaker & firearm & couch & table & cellphone & watercraft & mean \\
\hline
\multirow{7}{*}{CD$\downarrow$} & 3D-R2N2~\cite{choy20163d} &  0.895 & 1.891 & 0.735 & 0.845 & 1.432 & 1.707 & 4.009 & 1.507 & 0.993 & 1.135 & 1.116 & 1.137 & 1.215 & 1.445 \\
\cline{2-16}
& PSGN~\cite{fan2017point} & 0.430 & 0.629 & 0.439 & 0.333 & 0.645 & 0.722 & 1.193 & 0.756 & 0.423 & 0.549 & 0.517 & 0.438 & 0.633 & 0.593 \\
\cline{2-16}
& Pixel2Mesh~\cite{wang2018pixel2mesh} & 0.477 & 0.624 & 0.381 & 0.268 & 0.610 & 0.755 & 1.295 & 0.739 & 0.453 & 0.490 & 0.498 & 0.421 & 0.670 & 0.591 \\
\cline{2-16}
& GenRe~\cite{genRe} & 0.405 & 0.561 & 0.388 & 0.263 & 0.592 & 0.708 & 1.207 & 0.681 & 0.377 & 0.452 & 0.439 & 0.365 & 0.592 & 0.541 \\
\cline{2-16}
& GAL~\cite{jiang2018gal}  & 0.379 & 0.526 & 0.404 & 0.265 & 0.544 & 0.703 & 1.134 & 0.689 & 0.451 & 0.374  & 0.415 & 0.360 & 0.578 & 0.525 \\
\cline{2-16}
& PCDNet~\cite{nguyen2019graphx} & 0.116 & 0.189 & 0.265 & \textbf{0.184} & 0.306 & 0.248 & 0.523 & 0.419 & 0.119 & 0.254 & 0.284 & 0.155 & 0.210 & 0.252 \\
\cline{2-16}
& Ours    & \textbf{0.109} & \textbf{0.170} & \textbf{0.241} & 0.209 & \textbf{0.253} & \textbf{0.224} & \textbf{0.478} & \textbf{0.392} & \textbf{0.110} & \textbf{0.221} & \textbf{0.269} & \textbf{0.137} & \textbf{0.184} & \textbf{0.246}\\
\hline
\hline
\multirow{6}{*}{EMD$\downarrow$} & 3D-R2N2~\cite{choy20163d} &  0.606 & 1.136& 2.520 & 1.670 & 1.466 & 1.667 & 1.424 & 2.732 & 0.688 & 2.114 & 1.641 & 0.912 & 0.935 & 1.501 \\
\cline{2-16}
& PSGN~\cite{fan2017point} &  0.396 & 1.113 & 2.986 & 1.747 & 1.946 & 1.891 & 1.222 & 3.490 & 0.397 & 2.207 & 2.121 & 1.019 & 0.945 & 1.653\\
\cline{2-16}
& Pixel2Mesh~\cite{wang2018pixel2mesh} & 0.579 & 0.965 & 2.563 & 1.297 & 1.399 & 1.536 & 1.314 & 2.951 & 0.667 & 1.642 & 1.480 & 0.724 & 0.814 & 1.380 \\
\cline{2-16}
& GAL~\cite{jiang2018gal}  &  0.497 & 0.854 & 2.543 & 1.288 & 1.286 & 1.501 & 1.209 & 2.845 & 0.662 & 1.489 & 1.377 & 0.631 & 0.702 & 1.298 \\
\cline{2-16}
& PCDNet~\cite{nguyen2019graphx} &  0.167 & 0.253  & 0.414 & 0.354 & 0.389 & 0.295 & 0.476 & 0.528 & \textbf{0.132} & 0.386 & 0.412 & 0.201 & 0.243 & 0.337 \\
\cline{2-16}
& Ours    &  \textbf{0.156} & \textbf{0.244} & \textbf{0.340} &\textbf{ 0.341} & \textbf{0.334} & \textbf{0.273} & \textbf{0.415} & \textbf{0.517} & 0.139 & \textbf{0.319} & \textbf{0.364} &\textbf{ 0.196} &\textbf{ 0.226} &\textbf{ 0.305} \\
\hline
\end{tabular}
\end{adjustbox}
\end{center}
\end{table*}
%%%%

%%%%

\section{Experiments}

%\textbf{Dataset:} We train and evaluate the proposed networks using the ShapeNet dataset \cite{chang2015shapenet}, which contains a large collection of categorized 3D CAD models. For fair comparison, we use the same training/testing split as in Choy et. al. \cite{choy20163d}.

\textbf{Training Details:} Our networks are optimized using the Adam optimizer. To initialize our networks properly, we follow a two-stage training procedure: the depth estimation network and the point cloud completion network are first pretrained separately to predict the depth maps and the complete point clouds. The depth estimation network is trained with the L2 loss. Note that the ground truth depth map is the only ground truth we need to supervise the depth estimation. The partial point cloud is obtained from the depth map using the camera model (pure geometry transformation), so the full supervision for the partial point cloud is also the ground truth depth map, which is already used in the pipeline.
For pre-training the point cloud completion network, the ground-truth full point cloud is used as target and penalised by the Chamfer distance loss. This is how the network infers what to fill in for the missing parts of 3D point cloud.
Then the self-supervisions from Eq.~\ref{eq:2} and Eq.~\ref{eq:3} are used as complementary constraints in the joint end-to-end training. We also tried to use the ground truth full point cloud to supervise the partial point cloud, but the results are similar. However, when applying/fine-tuning the model to other real-world datasets without 3D ground truth, the self-supervision defined by Eq.~\ref{eq:2} can be used to regularize the partial point cloud to be consistent with the predicted full point cloud.

%%%%
\begin{table}[t]
\caption{The IoU of the 3D reconstruction results on ShapeNet. It is shown that our proposed method achieves higher IoU for most of the categories and a higher overall IoU}
\label{table:IoU}
\begin{center}
\begin{adjustbox}{width=0.45\textwidth}
\begin{tabular}{c|c|c|c|c|c|c|c}
\hline
  & \multicolumn{3}{c|}{3D-R2N2} & \multirow{2}{*}{PSGN} & \multirow{2}{*}{GAL} & \multirow{2}{*}{PCDNet} & \multirow{2}{*}{Ours} \\
\cline{2-4}
  & 1 view  & 3 views & 5 views  &  &  &  &  \\
\hline
\hline
airplane   & 0.513 & 0.549 & 0.561 & 0.601 &  0.685  & \textbf{0.758}&  0.682\\
bench   & 0.421 & 0.502 & 0.527 & 0.550 & 0.709 &\textbf{ 0.725} & 0.713 \\
cabinet  & 0.716 &  0.763 & 0.772 & 0.771 & 0.772 & 0.770 &\textbf{0.809} \\
car   & 0.798 & 0.829 & \textbf{0.836}& 0.831 & 0.737 & 0.819 &0.725 \\
chair   & 0.466 & 0.533 & 0.550 & 0.544 & 0.700  & 0.663 &\textbf{0.702} \\
monitor  & 0.468 & 0.545 & 0.565& 0.552 & 0.804  & 0.735  &\textbf{0.819} \\
lamp   & 0.381 & 0.415  & 0.421 & 0.462 &  0.670  & 0.516 &\textbf{0.674}\\
speaker   & 0.662 & 0.708 & 0.717 & 0.737& 0.698 & 0.708 & \textbf{0.743} \\
firearm  & 0.544 & 0.593 &0.600 & 0.604 &  0.715 & 0.747 &\textbf{0.753}\\
couch    & 0.628 & 0.690 &0.706 & 0.708  &  0.739 & \textbf{ 0.770}&0.752\\
table     & 0.513 & 0.564 &0.580  & 0.606 & 0.714&  0.605&\textbf{0.725} \\
cellphone    & 0.661 & 0.732&0.754  & 0.749 &  0.773  & \textbf{0.857} &0.789 \\
watercraft   & 0.513 &  0.596 & 0.610& 0.611 & 0.675  & \textbf{0.754} & 0.677 \\
\hline
mean    &0.560 & 0.617 &0.631 & 0.640  &  0.712  & 0.725 & \textbf{0.736} \\
\hline
\end{tabular}
\end{adjustbox}
\end{center}
\end{table}

%%%%%
\begin{figure*}[t]
\includegraphics[width=0.95\textwidth]{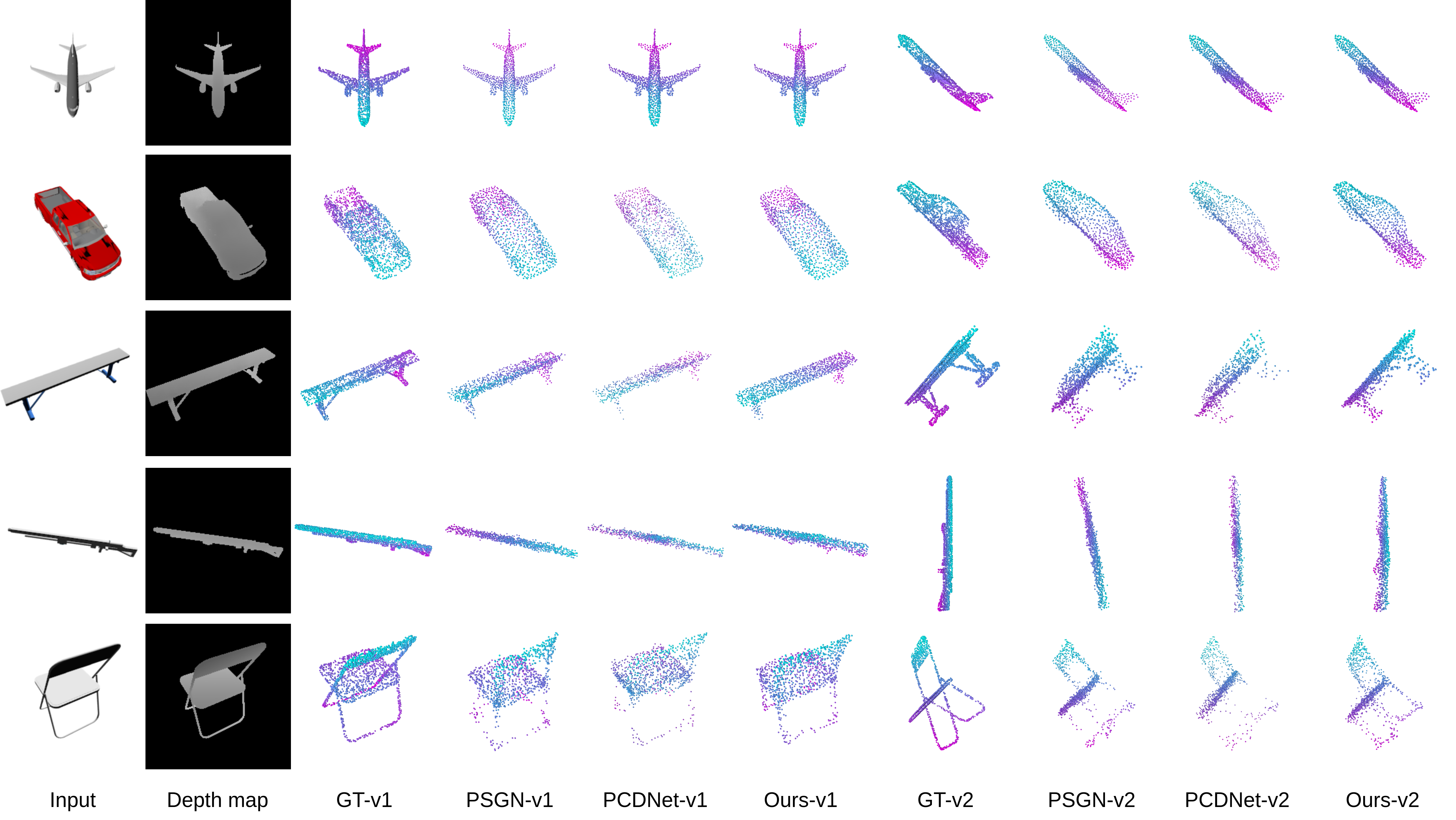}
\centering
\caption{Qualitative results for the ShapeNet dataset. We demonstrate the reconstruction results from two representative viewpoints $v1$ and $v2$. Compared to PSGN and GAL, the proposed method is better in capturing the overall shape and in generating finer details. }
\label{fig:results}
\end{figure*}

%%%%%

\textbf{Evaluation Metric:} We evaluate the different methods using three metrics: point-cloud based Chamfer Distance (CD), point-cloud based Earth Mover's Distance (EMD) and voxel-based Intersection over Union (IoU). 

The Chamfer Distance measures the distance between the predicted point cloud $P_p$ and the ground truth point cloud $P_{gt}$. This loss is defined by: 
\begin{equation} 
    \begin{aligned}
    L_{CD}(P_p,P_{gt})=
    & \frac{1}{|P_p|}\sum_{x\in P_p}\underset{y\in P_{gt}}{min}\|x-y\|^2_2 + \\
    & \frac{1}{|P_{gt}|}\sum_{y\in P_{gt}}\underset{x\in P_p}{min}\|x-y\|^2_2
    \end{aligned}
\end{equation}

The Earth Mover's Distance requires $P_p, P_{gt} \subseteq {R}^3$ to have equal size $s=|P_p|=|P_{gt}|$. The EMD distance is defined by:
\begin{equation} \label{eq:5}
    L_{EMD}(P_p, P_{gt}) = 
    \frac{1}{|s|}\underset{\phi: P_p \to P_{gt}}{min}
    \sum_{x\in P_p}\|x-\phi(x)\|^2_2
\end{equation}
where $\phi: P_p\to P_{gt}$ is a bijection. A lower CD/EMD value represents a better reconstruction result.

To compute the IoU of the predicted and ground truth point clouds, we follow the setting of GAL \cite{jiang2018gal}. Each point set is voxelized by distributing points on $32\times32\times32$ grids. The point grid for each point is defined as a $1\times1\times1$ grid centered at this point. For each voxel, the maximum intersecting volume ratio of each point grid and this voxel is calculated as the occupancy probability. IoU is defined as follows: 
\begin{equation} \label{eq:6}
    IoU=\frac{\sum_i\mathbbm{1}[V_{gt}(i)V_p(i)>0]}
    {\sum_i\mathbbm{1}[V_{gt}(i) + V_p(i)>0]}
\end{equation}
where $V_{gt}$ and $V_p$ are the voxelized ground-truth and prediction, respectively. $i$ is the index of the voxels. $\mathbbm{1}$ is an indicator function. A higher IoU value indicates a better point cloud prediction.

%%%%
\begin{table}[t]
\caption{Verification of the depth estimation module. The performance of the VGG-16-based network is similar to FCRN. Therefore we choose VGG-16 for simplification. The depth estimation network strongly benefits from the 3D self-supervision approach of the 3D-2D refinement module. All numbers are scaled by a factor of 10}
\label{table:depth estimation}
\begin{center}
\begin{adjustbox}{width=0.4\textwidth}
\begin{tabular}{c|c|c|c}
\hline
  & \multirow{2}{*}{FCRN}   & \multicolumn{2}{c}{Our depth module (VGG-16)}\\
\cline{3-4}
  &   &  w/o refinement &  w/ refinement  \\
\hline
airplane  & 0.152 & 0.166 & \textbf{0.105}\\
bench    &  0.424 & 0.421 & \textbf{0.358}\\
cabinet   & 0.576  & 0.584 & \textbf{0.499}\\
car    &  0.273 & 0.267 & \textbf{0.258}\\
chair   &  0.926  & 0.968 &\textbf{0.890}\\
lamp    &  0.417 & 0.428  & \textbf{0.399}\\
monitor   &  0.684 & 0.707 & \textbf{0.639}\\
rifle   & 0.051 & 0.047  & \textbf{0.046}\\
sofa    &  0.554  & 0.551 & \textbf{0.497}\\
speaker    &  0.741 & 0.731 & \textbf{0.672}\\
table      &  0.287 & 0.298  & \textbf{0.282}\\
telephone    & 0.261   & 0.259& \textbf{0.237} \\
vessel     & 0.270 & 0.271 & \textbf{0.260}\\
\hline
\hline
mean     & 0.432  &0.438 &\textbf{0.395} \\
\hline
\end{tabular}
\end{adjustbox}
\end{center}
\end{table}

%%%%

\textbf{ShapeNet Dataset:} We train and evaluate the proposed networks using the ShapeNet dataset \cite{chang2015shapenet} containing a large collection of categorized 3D CAD models. The same training/testing split as in 3D-R2N2 \cite{choy20163d} is used. Since the proposed method needs the ground truth depth maps to guide the depth intermediation step, we re-render the $RGB$ images and the corresponding depth maps for each instance from 12 different views. For a fair comparison, we re-produce the results for GenRe \cite{genRe}, GAL \cite{jiang2018gal} and show the quantitative comparison of CD and EMD metric in Table \ref{table:overall}.

3D-R2N2 \cite{choy20163d} takes as an input one or more images of an object which are taken from different viewpoints. The method outputs a reconstruction of the object in the form of a 3D occupancy grid. PSGN \cite{fan2017point} utilizes fully-connected layers and deconvolutional layers to predict 3D points directly from 2D images. Pixel2Mesh \cite{wang2018pixel2mesh} designs a projection layer which incorporates perceptual image features into 3D geometry represented by graph based convolutional network. It predicts 3D geometry in a coarse to fine fashion and generates a 3D mesh model from a single RGB image. GenRe \cite{genRe} combines 2.5D representations of visible surfaces, spherical shape representations of both visible and non-visible surfaces and 3D voxel-based representations, in a principled manner to capture generic shape priors. GAL \cite{jiang2018gal} proposes a complementary loss, the geometric adversarial loss, to geometrically regularize predictions from a global perspective. PCDNet \cite{nguyen2019graphx} deforms a random point set according to an input object image and produce a point cloud of the object by a network consisting of GraphX. As shown in Table \ref{table:overall}, our method outperforms existing methods for most of the categories for both CD and EMD metric. In addition, our method achieves a lower overall mean score.

%%%%
%\begin{table}[t]
%\caption{Quantitative comparison of CD and EMD metric on the NED dataset. Both CD and EMD numbers are scaled by a factor of 100.}
%\label{table:NED}
%\begin{center}
%\begin{adjustbox}{width=0.48\textwidth}
%\begin{tabular}{c | c | c | c || c | c | c || c | c | c}
%\hline
% & \multicolumn{3}{c||}{CD} & \multicolumn{3}{c||}{EMD} & \multicolumn{3}{c}{IoU} \\
%\cline{2-10} 
%& PSGN & GAL & Ours & PSGN & GAL & Ours & PSGN & GAL & Ours \\
%\hline
%\hline
%hedges & 6.445 & 3.809 & \textbf{3.298} & 11.564 & 5.836 & \textbf{5.489}  & 0.526 & \textbf{ 0.704}  & 0.697 \\

%rocks & 4.592  & 2.775 &\textbf{ 2.481}  & 6.973 & 4.179 &\textbf{ 3.927}  & 0.383 & \textbf{0.596}   & \textbf{0.596}   \\

%topiaries & 3.702  & 2.285 &\textbf{ 2.065 } & 7.355 & 3.763  & \textbf{3.393}   & 0.435 & 0.633 &  \textbf{0.648 }  \\
%\hline
%mean  & 4.913 & 2.956 & \textbf{2.615}  & 8.631 & 4.593 & \textbf{4.270 } & 0.448 & 0.644 & \textbf{0.647}  \\
%\hline

%\end{tabular}
%\end{adjustbox}
%\end{center}
%\end{table}
%%%%

%%%%%
\begin{figure}[t]
\includegraphics[width=0.47\textwidth]{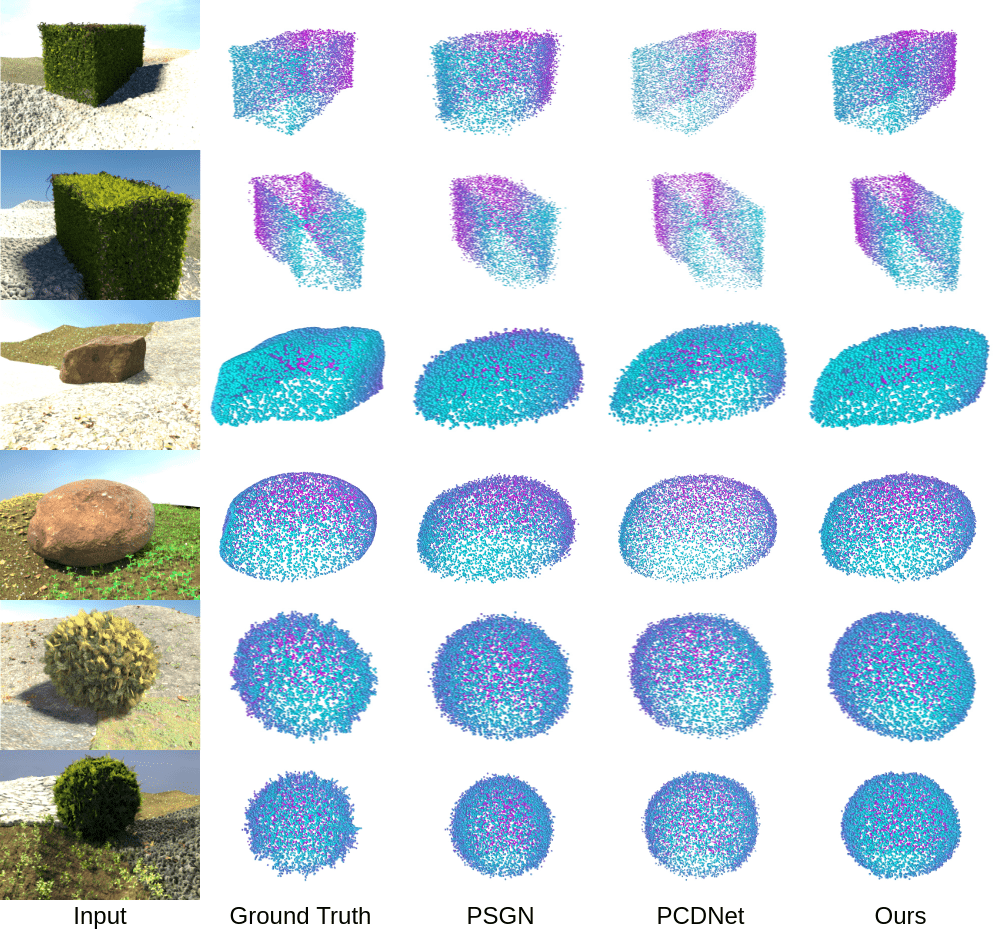}
\centering
\caption{Qualitative results on the object-centric NED dataset. Since in this setting the objects are relatively simple and regular, both GAL and our method can generate accurate 3D point clouds while PSGN fails for some parts of the 3D shapes. }
\label{fig:NED}
\end{figure}

%%%%%

%%%%
\begin{table}[t]
\caption{Quantitative comparison on the NED dataset. Our proposed method outperforms the other methods to recover the point clouds of the three categories of NED }
\label{table:NED}
\begin{center}
\begin{adjustbox}{width=0.35\textwidth}
\begin{tabular}{c | c | c | c | c }
\hline
\multicolumn{2}{c|}{} & CD$\downarrow$ & EMD$\downarrow$ & IoU$\uparrow$ \\
\hline
\hline
\multirow{3}{*}{hedge} & PSGN & 0.645& 1.156& 0.526\\
                       & PCDNet  & 0.311& 0.484& \textbf{0.704}\\
                       & Ours & \textbf{0.274}& \textbf{0.428}& 0.697\\
\hline
\multirow{3}{*}{rock} & PSGN & 0.459& 0.697& 0.583\\
                       & PCDNet  & 0.253& 0.396& 0.607\\
                       & Ours & \textbf{0.219}& \textbf{0.375}& \textbf{0.649}\\
\hline
\multirow{3}{*}{topiary} & PSGN & 0.370 & 0.736 & 0.435\\
                       & PCDNet  & 0.229& 0.376& 0.633\\
                       & Ours & \textbf{0.207}& \textbf{0.339}& \textbf{0.648}\\
\hline
\hline
\multirow{3}{*}{mean} & PSGN & 0.491& 0.863& 0.514\\
                       & PCDNet  & 0.264& 0.419& 0.648\\
                       & Ours & \textbf{0.233}& \textbf{0.381}&\textbf{0.665}\\
\hline
\end{tabular}
\end{adjustbox}
\end{center}
\end{table}
%%%%

A number of qualitative results are shown in Fig.~\ref{fig:results}. The first row shows that PSGN, PCDNet and our method perform well in generating the full point clouds for some simple objects and regular shapes. In the second and third row, our method provides accurate structures, while either PSGN or PCDNet fail at recovering parts of the 3D shapes (e.g. the rear end of the Pick-up in the second row, the backrest of the bench in the third row). It is shown that our method also generates a better pose estimation, see viewpoint $v2$ in the fourth row. Further, the result of our proposed method is more aligned with the ground truth than PSGN. Failure cases are shown in the last row which all methods are not able to capture the correct structure of the chair leg.

Table \ref{table:IoU} shows the IoU value for each category in ShapeNet dataset. It can be derived that our method obtains a better IoU for most of the categories. Our method explicitly incorporates the camera model as a geometrical constraint to regulate the 2D-3D domain transfer. As a consequence, the generated point clouds are more aligned with the ground truth point clouds. 

We also verify the choice of the depth estimation network and the benefit from the 3D-2D refinement module. FCRN \cite{fcrn} is a very deep depth estimation network based on ResNet-50. Since the object-level depth estimation in our task is relatively simple, the performance of FCRN is similar to the shallow VGG-16-based architecture, as shown in the first two columns in Table \ref{table:depth estimation}. Therefore, in our depth intermediation module, we choose VGG-16 for simplification. The third column of Table \ref{table:depth estimation} shows that the depth estimation network benefits significantly from the 3D self-supervision strategy. As shown in Fig.~\ref{fig:partial point cloud}, the depth estimation with only 2D supervision may suffer from the estimation error near the boundaries along the frustum. With our 3D-2D refinement, the generated full point cloud is utilized as 3D self-supervision to reduce the estimation error.

%%%%%
\begin{figure}[t]
\includegraphics[width=0.45\textwidth]{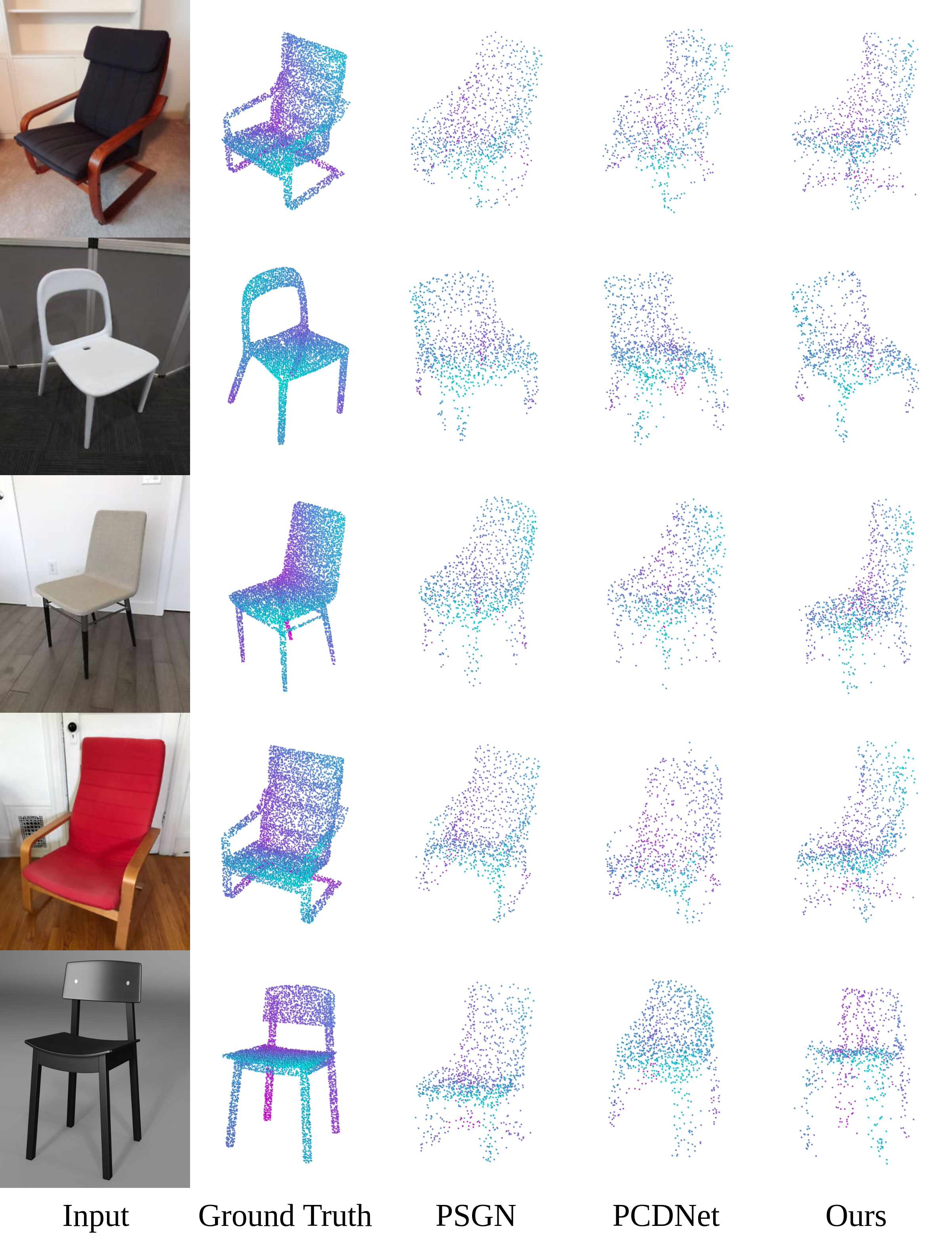}
\centering
\caption{Qualitative results on chair subset of Pix3D dataset. Since in this setting the task is relatively challenging, all three methods perform reasonable well in visually perspective. But our method can capture more details of the shapes.}
\label{fig:pix3d}
\end{figure}

%%%%%

\textbf{NED Dataset:} We consider the The Natural Environment Dataset (NED) \cite{NED} to evaluate our proposed pipeline. In contrast to man-made objects, the NED dataset consists of (3D) synthetic scene-centric images from outdoor (natural) environments like gardens and parks. Images are rendered with the physics-based Blender Cycles engine\footnote{https://www.blender.org/}. The model textures and skies are used from real-world images to provide a realistic look of the scenes. 
Three categories are selected: hedges, rocks and topiaries. 
We follow the same rendering (scene-centric) settings of the dataset to render the object-centric images. 
%The training set consists of around 50 instances for each category and for each instance we render from 12 different views. 
We train and test PSGN, PCDNet and our method on these images, see Fig.~\ref{fig:NED}. Since in this setting the objects are relatively simple and regular, both PCDNet and our method can generate accurate 3D point clouds while PSGN fails for some parts of the 3D shapes (for example the wrong oval shape of the hedges in the first two rows and the missing finer shape details of the rock in the third row for PSGN). Table~\ref{table:NED} shows the quantitative results for this dataset. Our proposed method outperforms the other methods to recover the point clouds of the three categories of NED.

%%%%%
\begin{figure}[t]
\includegraphics[width=0.45\textwidth]{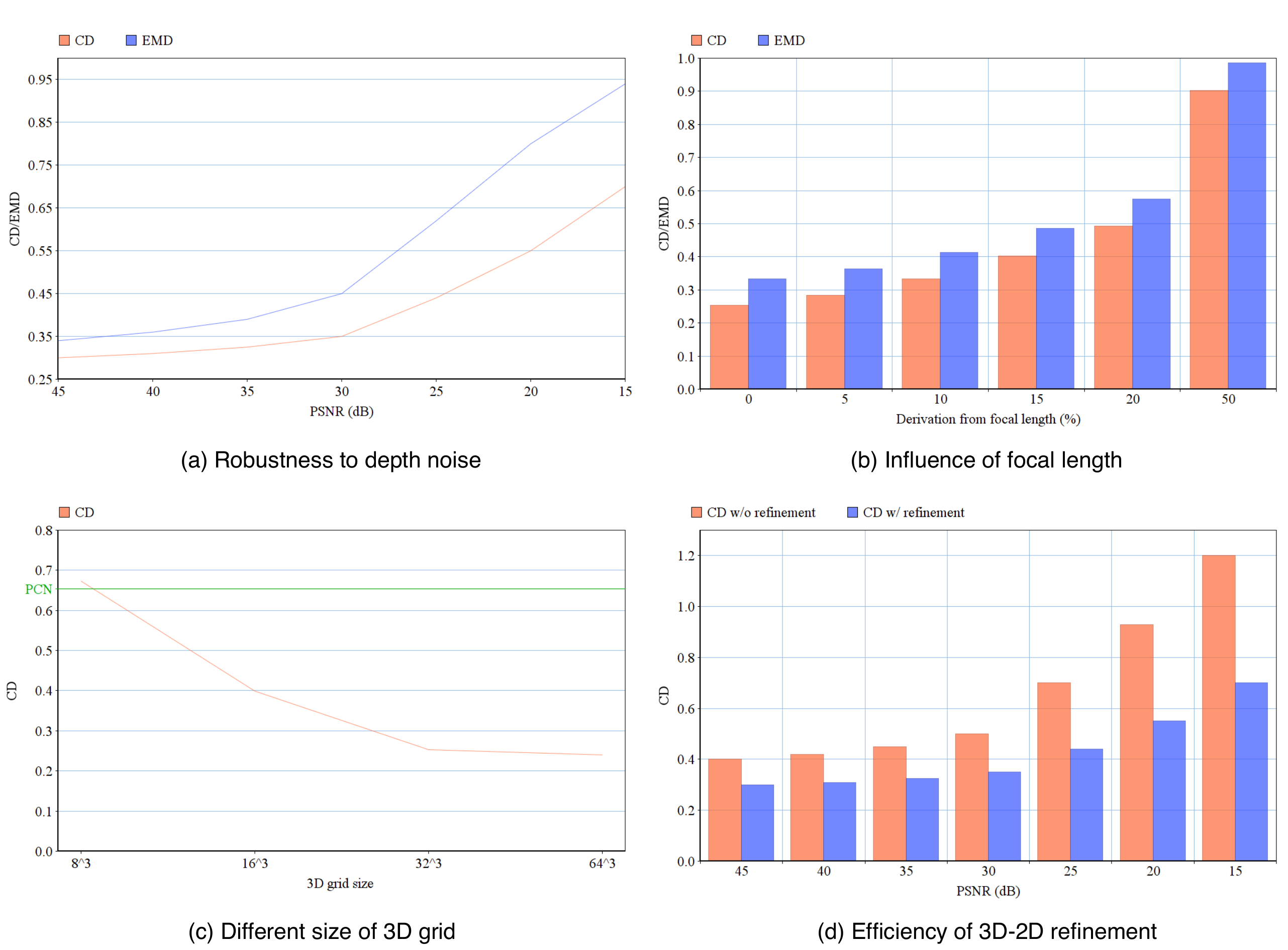}
\centering
\caption{Ablation study for the different components of the proposed pipeline.}
\label{fig:ablation_study}
\end{figure}
%%%%%

%%%%
\begin{table}[t]
\caption{Quantitative comparison of CD and EMD metric for the chair subset of the Pix3D dataset. The proposed method outperforms the other state-of-the-art methods}
\label{table:Pix3D}
\begin{center}
\begin{adjustbox}{width=0.4\textwidth}
\begin{tabular}{c | c | c | c || c | c | c  }
\hline
 & \multicolumn{3}{c||}{w/o fine-tuning} & \multicolumn{3}{c}{w/ fine-tuning} \\
\cline{2-7} 
& CD$\downarrow$ & EMD$\downarrow$ & IoU$\uparrow$ & CD$\downarrow$ & EMD$\downarrow$ & IoU$\uparrow$   \\
\hline
\hline
PSGN & 0.389 & 0.453 & 0.143 & 0.357 & 0.412 & 0.167 \\

PCDNet & 0.297  & 0.386 & 0.148 & 0.261 & 0.354 & 0.185 \\

Ours & \textbf{0.193} & \textbf{0.249} &\textbf{0.168} &\textbf{0.142} & \textbf{0.213} & \textbf{0.244} \\
\hline

\end{tabular}
\end{adjustbox}
\end{center}
\end{table}
%%%%

\textbf{Pix3D Dataset:} Pix3D \cite{pix3d} is a large-scale dataset containing diverse image-shape pairs with pixel-level 2D-3D alignment. For a fair comparison, the chair subset is selected. The chair subset of Pix3D \cite{pix3d} contains 3839 images with the corresponding 3d models. To fine-tune the models trained on ShapeNet, the first 3000 image-shape pairs are used as training data. The last 839 pairs are testing data for both models without and with fine-tuning. Fig.~\ref{fig:pix3d} demonstrates a number of quantitative results from fine-tuning models for this dataset. Since in this setting the task is relatively challenging, all three methods perform reasonable well in visually perspective. But our method can capture more details of the shapes (for example the chair leg in the first row, the pose of the chair in the fourth row and the overall shape of the chair in the last row). Table \ref{table:Pix3D} shows the results without and with fine-tuning for each method. It can be derived that for both cases, the proposed method outperforms the other state-of-the-art methods.

%%%%%
%\begin{figure}[t]
%\includegraphics[scale=0.28]{latex/depth_noise.pdf}
%\centering
%\caption{Robustness to depth noise. }
%\label{fig:depth_noise}
%\end{figure}
%%%%%

%%%%%
%\begin{figure}[t]
%\includegraphics[scale=0.28]{latex/focal_length.pdf}
%\centering
%\caption{Influence of focal length. }
%\label{fig:focal_length}
%\end{figure}
%%%%%

%%%%%
%\begin{figure}[t]
%\includegraphics[scale=0.28]{latex/refinement.pdf}
%\centering
%\caption{Efficiency of 3D-2D refinement}
%\label{fig:refinement}
%\end{figure}
%%%%%

\textbf{Ablation Study:} In this section, ablation experiments are conducted to analyze the performance of different components in our full pipeline. To this end, the chair subset of ShapeNet is selected to re-train the proposed method.

\begin{table}[t]
\caption{The performance gap with and without 3D-2D refinement module comparing to baseline PSGN (No adding Gaussian noise)}
\label{table:refinement}
\begin{center}
\begin{adjustbox}{width=0.38\textwidth}
\begin{tabular}{l|c|c}
\hline
  &  CD$\downarrow$  &  IoU$\uparrow$  \\
\hline
PSGN (Baseline)    &   0.645  & 0.544  \\

Ours (w/o 3D-2D refinement) &  0.485 &  0.626 \\

Ours (w/ 3D-2D refinement)   &   \textbf{0.253}  & \textbf{0.702}   \\
\hline
\end{tabular}
\end{adjustbox}
\end{center}
\end{table}

\emph{Depth intermediation component:} An important component of our approach is the depth intermediation module which regulates the 2D-3D domain transfer. To test the influence of the quality of the depth map estimation, during evaluation, different levels of Gaussian noise are added to the predicted depth maps to verify the robustness of the proposed method to depth noise. PSNR is used to measure the amount of noise. A lower PSNR value indicates a noisier image. In general, for computer vision tasks, acceptable values for PSNR are considered to be above 30dB. As shown in Fig.~\ref{fig:ablation_study} (a), our proposed method is quite robust in the range above 30dB. 

\emph{Camera model component:} Another component is the camera model which is used as a geometrical constraint to steer the 2D-3D domain transfer. We assume that the focal length is known (which is not always the case). Therefore, in the experiments, we analyze the robustness of the used camera model for different focal length estimations in terms of deviations from the ground truth (focal length). Fig.~\ref{fig:ablation_study} (b) shows the CD and EMD with regard to the deviations from the ground truth focal length. Our proposed method can still provide reasonable results even when the estimated focal lengths are 20\% off from the ground truth focal length.

\emph{3D grid basis point set component:} The 3D grid basis point set is used to encode the unordered partial point cloud to learn context-aware features. To verify the influence of the size of the 3D grid, we train models with different sizes of 3D grid basis point set. The baseline is the PCN \cite{yuan2018pcn} without any 3D grid encoding. As shown in Fig.~\ref{fig:ablation_study} (c), as the 3D grid size increases, the performance of the network is also improved. To achieve a balance between the effect and efficiency, we choose the 3D grid size as $32^3$ in this work.

\emph{3D-2D refinement component:} The 3D-2D refinement module in our proposed pipeline is crucial to reduce the depth estimation errors. In Table  \ref{table:refinement} we show the performance gap with and without 3D-2D refinement module comparing to baseline PSGN on the chair subset of ShapeNet. In order to verify the robustness of the 3D-2D refinement module against noise, we train models for different Gaussian noise levels. Here injecting Gaussian noise to alter the depth predictions is to simulate the situations that the depth predictions are inaccurate. As shown in Fig.~\ref{fig:ablation_study} (d), the performance gap enlarges dramatically when PSNR decreases, which shows the robustness of the proposed 3D-2D refinement module with respect to the inaccurate depth estimation. This indicates that our 3D-2D refinement module can greatly reduces the depth noise and produces more accurate point clouds.

%%%%%
\begin{figure}[t]
\includegraphics[width=0.45\textwidth]{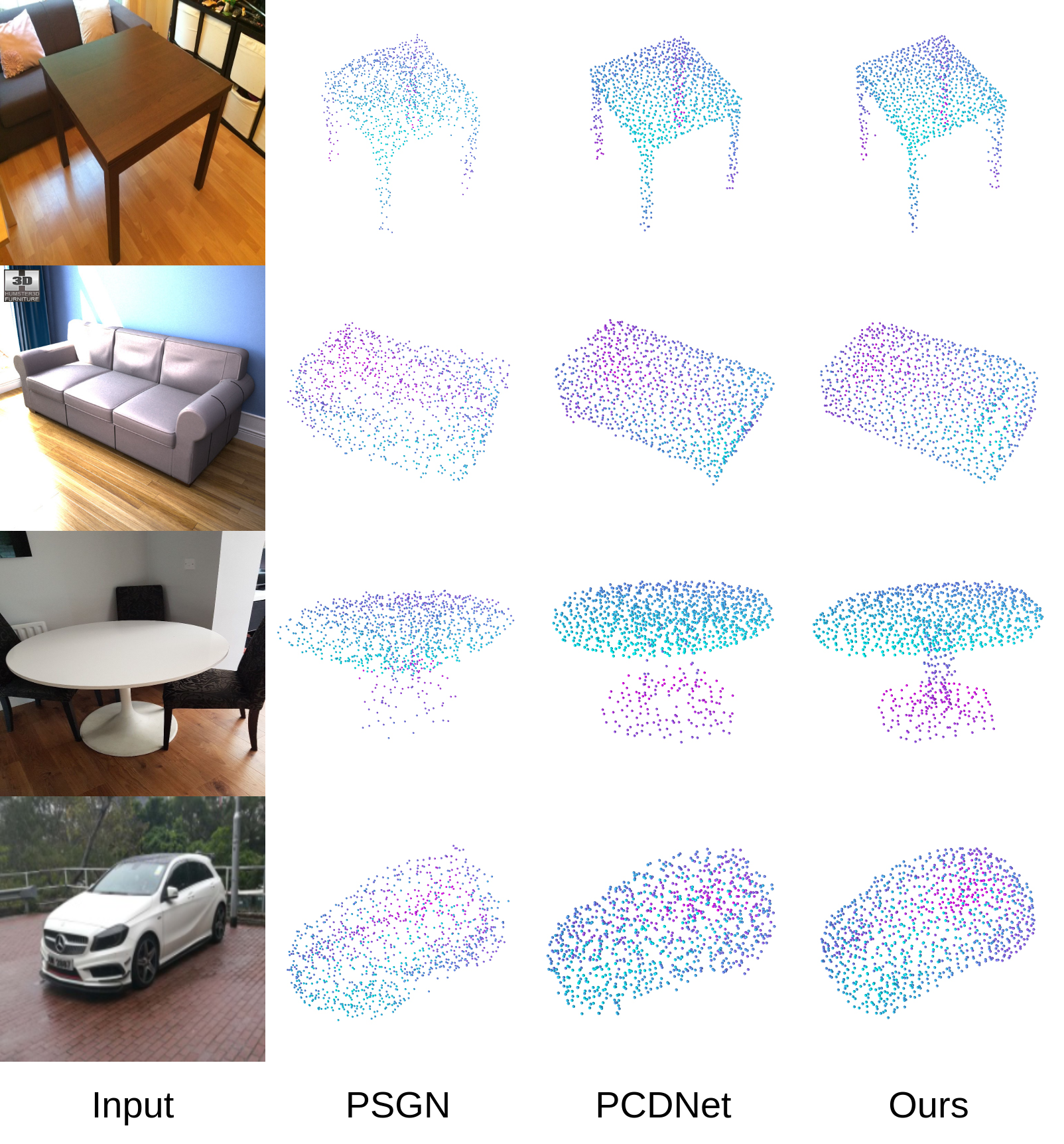}
\centering
\caption{Qualitative results for a number of real-world images. Our proposed method (trained on synthetic data) generalizes well to real-world images.}
\label{fig:real-world}
\end{figure}

%%%%%

\textbf{Images In The Wild:} We also test the generalizability of our approach on real-world images. We use the model trained on the ShapeNet dataset and directly run it on real-world images which are randomly selected from the Internet (with manually created masks). We consider these real-world images as captured by different cameras and with different camera parameters. We use estimated focal lengths during evaluation. Results are shown in Fig.~\ref{fig:real-world}. Our proposed method can generate overall smooth point clouds (e.g. the second and fourth row) and capture more details (table leg in the third row) for the objects in the in-the-wild images. It indicates that our model trained on synthetic data generalizes well to the real-world images.

%-------------------------------------------------------------------------
\section{Conclusion}

In this paper, we propose an efficient framework to generate 3D point clouds from single monocular $RGB$ images by sequentially predicting the depth maps and inferring the complete 3D object shapes. Depth estimation and camera model are explicitly incorporated in our pipeline as geometrical constraints during both training and inference. We also enforce the alignment between the predicted full 3D point clouds and the corresponding estimated depth maps to jointly optimize both depth intermediation and the point completion module. 

Both qualitative and quantitative results on ShapeNet, NED and Pix3D show that our method outperforms existing methods. Furthermore, it also generates precise point clouds for real-world images. In the future, we plan to extend our framework to scene-level point cloud generation.

{\small
\bibliographystyle{ieee_fullname}
\bibliography{egbib}
}

\end{document}